\documentclass[letterpaper, 10 pt, conference]{ieeeconf}  

\IEEEoverridecommandlockouts                              

\usepackage{soul}
\usepackage{graphicx}
\usepackage{tabularx}
\usepackage[table]{xcolor}
\usepackage{colortbl}
\usepackage{cite}

\title{\LARGE \bf
A Deep Learning-Driven Autonomous System for Retinal Vein Cannulation: Validation Using a Chicken Embryo Model}

\author{Yi Wang$^{1,*}$, Peiyao Zhang$^{2,*}$, Mojtaba Esfandiari$^{2}$, \IEEEmembership{Graduate Student Member, IEEE}, \\ Peter Gehlbach$^{3}$,  {\it Member, IEEE},  Iulian I. Iordachita$^{2}$, \IEEEmembership{Senior Member, IEEE} 
\thanks{$*$ Equal Contribution}
\thanks{$^{1}$ Yi Wang is with the Department of Robotics and also the Laboratory for Computational Sensing and Robotics at the Johns Hopkins University, Baltimore, MD, 21218, USA.
       (e-mail: {\tt\small  ywang779@jhu.edu})}%
\thanks{$^{2}$ Peiyao Zhang, Mojtaba Esfandiari, and Iulian Iordachita are with the Department of Mechanical Engineering and also the Laboratory for Computational Sensing and Robotics at the Johns Hopkins University, Baltimore, MD, 21218, USA.
       (e-mail: {\tt\small  pzhang24, mesfand2, iordachita@jhu.edu})}%
\thanks{$^{3}$ Peter Gehlbach is with the Wilmer Eye Institute, Johns Hopkins Hospital, Baltimore, MD, 21287, USA.
       (e-mail: {\tt\small pgelbach@jhmi.edu})}%
\thanks{This work is supported by the U.S. National Institutes of Health under the grants numbers R01EB023943, R01EB025883, R01EB034397, and partially by JHU internal funds.}
}

\begin{document}

\maketitle
\thispagestyle{empty}
\pagestyle{empty}

\begin{abstract}
Retinal vein cannulation (RVC) is a minimally invasive microsurgical procedure for treating retinal vein occlusion (RVO), a leading cause of vision impairment. However, the small size and fragility of retinal veins, coupled with the need for high-precision, tremor-free needle manipulation, create significant technical challenges. These limitations highlight the need for robotic assistance to improve accuracy and stability.
This study presents an automated robotic system with a top-down microscope and B-scan optical coherence tomography (OCT) imaging for precise depth sensing. Deep learning-based models enable real-time needle navigation, contact detection, and vein puncture recognition, using a chicken embryo model as a surrogate for human retinal veins. The system autonomously detects needle position and puncture events with 85\% accuracy. The experiments demonstrate notable reductions in navigation and puncture times compared to manual methods. Our results demonstrate the potential of integrating advanced imaging and deep learning to automate microsurgical tasks, providing a pathway for safer and more reliable RVC procedures with enhanced precision and reproducibility.

\end{abstract}

\section{INTRODUCTION} \label{sec:Introduction}

\begin{figure}[t]  
    \centering
    \includegraphics[width=\linewidth]{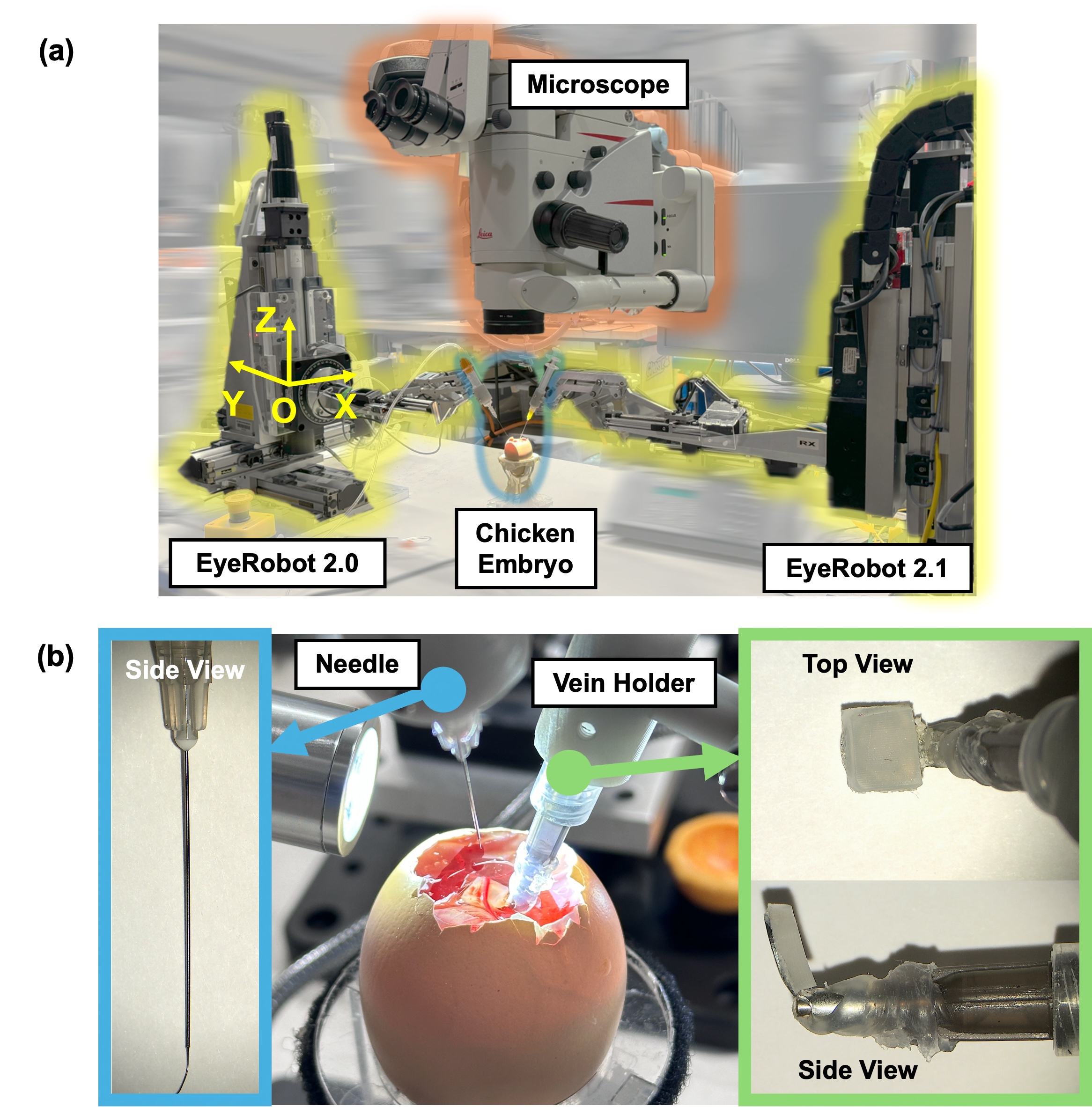} 
    \caption{Experimental setup and key components of the autonomous system for retinal vein cannulation. (a) The overall setup shows EyeRobot 2.0 and EyeRobot 2.1, which work in tandem during the procedure. EyeRobot 2.0 is responsible for needle manipulation, while EyeRobot 2.1 stabilizes the target vein (diameter approximately 300-400 micrometers) in the chicken embryo. Unlike EyeRobot 2.0, which operates autonomously during needle navigation and puncture, EyeRobot 2.1 is manually positioned before the procedure to ensure proper alignment with the target vein. The holding mechanism in EyeRobot 2.1 secures the vein in place without active control during the procedure, preventing unnecessary movement that could interfere with the puncture process
(b) Close-up images of the surgical needle and vein holder. The side view of the needle and the top and side view of the vein holder demonstrate the orientation and structure used during the experiments. }
    \label{fig:1}
    \vspace{-12pt}
\end{figure}

Retinal vein occlusion (RVO) occurs due to the blockage of a retinal vein by a thrombus, leading to transient or permanent vision loss \cite{sivaprasad2015royal}. Current treatments focus on managing complications, but no standardized surgical approach exists for thrombus removal.
A 2015 meta-analysis identified RVO as the second most prevalent retinal vascular disease globally, affecting 28.06 million people aged 30-89, including 23.38 million branch RVO (BRVO) and 4.67 million central RVO (CRVO) \cite{song2019global}. Retinal vein cannulation (RVC) involves inserting a micro-needle into the occluded retinal vein, followed by injecting a thrombolytic agent to dissolve the clot \cite{tameesh2004retinal}. However, manual RVC is hindered by physiological tremors that introduce instability during needle manipulation.  The RMS amplitude of an ophthalmic surgeon's hand tremor is approximately 182 $\mu$m \cite{riviere2000study}, while the average diameter of retinal veins at their widest is 151.32 $\pm$ 15.22 $\mu$m \cite{goldenberg2013diameters}. Maintaining needle stability inside the vein for prolonged durations further exacerbates the difficulty due to tremors and fatigue \cite{willekens2017robot}.
To overcome these challenges, several surgical robotic systems have been developed, including the Steady Hand Eye Robot (SHER) \cite{uneri2010new, esfandiari2024cooperative}, Preceyes \cite{meenink2013robot}, a compact robot with a hybrid mechanism for ophthalmic surgery \cite{nasseri2013introduction}, and others \cite{ gijbels2013design, vander2020robotic}. Despite these advancements, robotic-assisted RVC still faces challenges, particularly in visual feedback. Binocular microscopes often lack the depth perception required at the micron scale, increasing the risk of double puncturing the vein and damaging retinal tissues.

Several methods have been studied to overcome the challenge of depth perception in robot-assisted retinal microsurgery. One approach is to rely on deep learning networks to predict depth from a single binocular image. Kim \textit{et al}. \cite{kim2021towards} predicted the depth of the needle tip on a silicone eye phantom using a convolutional neural network (CNN) based on the shadow cue from the needle shaft. They enhanced the model's safety using chance-constrained optimization \cite{zhang2021towards} and evaluated it on \textit{ex vivo} porcine eyes \cite{zhang2023autonomous}. Others have attempted to obtain accurate depth information using OCT, a non-invasive method to acquire cross-sectional images of biological tissues.
The use of OCT images has been shown to enable automated subretinal injections with micrometer-level accuracy \cite{kim2024towards, zhang2024autonomous}, compensating both tissue deformation during tool-tissue interaction \cite{arikan2024real} and retina motion caused by respiration and heartbeat \cite{arikan2025towards, wu2025deep}. Our previous work \cite{zhang2024a} demonstrated that retinal vein cannulation can be achieved using keyboard teleoperation on intact porcine eyes.

In this work, we present a fully autonomous robotic system for retinal vein cannulation that integrates a top-down microscope with B-scan imaging. The B-scan images of the region of interest (ROI) provide precise depth information, enabling accurate evaluation of the interaction between the needle and the vein. The system leverages two EyeRobot platforms: one to hold and stabilize the vein, while the other manipulates the needle (see Fig. \ref{fig:1}). This configuration safeguards the needle during puncture and minimizes the risk of double puncture by maintaining vein stability. To achieve full autonomy, deep learning models were trained for needle tip detection, contact assessment, and puncture recognition, allowing the system to perform the entire procedure—from needle navigation to puncture detection—without manual intervention. While this system shows promising performance, several limitations remain. In particular, the current vision pipeline is sensitive to occlusion, motion artifacts, and lighting variations. These factors can reduce the visibility of the needle in microscope and B-scan images, which may lead to misclassification in contact and puncture stages. Moreover, since all models were trained on data collected from a single imaging setup and a single optical domain, the system may not generalize well to different imaging devices, tissue types, or lighting conditions without fine-tuning. These limitations suggest that future work should consider validating the models on external datasets or applying more generalizable architectures, such as self-supervised vision transformers or networks with geometry priors.


\section{MOTIVATION AND PROBLEM FORMULATION}
\subsection{Chicken Embryo Model}
The chicken embryo model is recognized for anatomical similarities to the human eye, particularly its transparent vascular system similar to human retinal veins.
These similarities make the chicken embryo (at around 12-14 days of hatching time) an excellent platform for simulating delicate ocular procedures, such as retinal vein cannulation \cite{patel2020comparison}.
By using the chicken embryo, surgical techniques can be refined and validated in an environment similar to the human retina.

\begin{figure}[ht]
    \includegraphics[width=\linewidth]{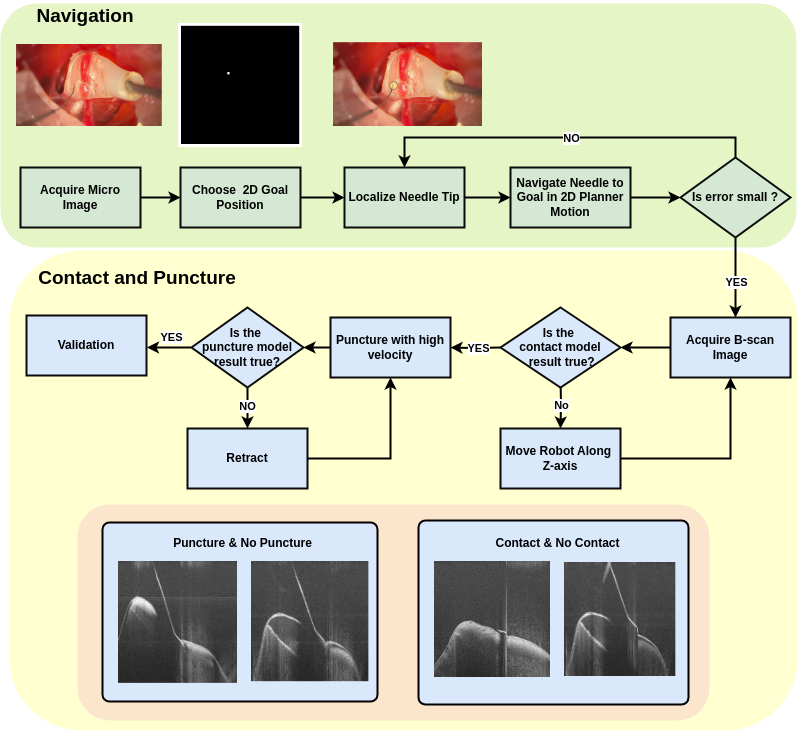} 

    \caption{Block diagram of the proposed workflow. The process is divided into two key phases: \textit{Navigation} and \textit{Contact and Puncture}. In the navigation phase, the surgeon selects a 2D goal position on a microscopic image, and the robot, using Detectron2 for needle tip detection, calculates the direction towards the goal position. Once the goal is reached, the system enters the \textit{Contact and Puncture} phase. A B-scan image is acquired to detect contact using ResNet18, followed by puncture detection using YOLOv8x. If no contact is detected, the needle is adjusted along the z-axis until contact is confirmed. The system then increases velocity for puncture and verifies the outcome using B-scan images. If no puncture is detected, the system retracts and reattempts the procedure. The procedure ends once a successful puncture is made.}
    \label{fig:2}
    \vspace{-12pt}
\end{figure}

\subsection{Motivation for Autonomous Microsurgery}

To address the challenges of manual RVC, including physiological tremors and limited real-time feedback, a transition to autonomous systems is proposed. Autonomous systems can eliminate human error, enhance precision, and ensure the fine control required. Incorporating such systems in RVC can reduce surgical failures and related complications.
An automated procedure will further ease the cognitive load on surgeons in such attention-intensive tasks. Overall, procedure acceptance and availability will increase, supporting future evaluation as a treatment for RVO.

\subsection{Deep Learning for Surgical Automation}
The effectiveness of this system depends on accurately interpreting high-resolution OCT B-scan images\cite{vander2020robotic}. Both microscope images and B-scan images can visualize micron-scale features of biological tissue. For instance, spectral-domain OCT can achieve an axial resolution of less than 5 $\mu m$ \cite{tao2010intraoperative}. 
Traditional image processing methods that rely on edge and intensity features often fail due to OCT variability and complexity.

However, deep learning models, particularly CNNs, offer a superior solution by learning complex patterns in the dataset automatically, which can make them well-suited for analyzing both microscope and B-scan OCT images.
 
Unlike rule-based algorithms that rely on predefined parameters, deep learning models can continuously improve their performance through exposure to large datasets \cite{bailly2022effects}. This ability to adapt and generalize supports precise, real-time decision-making needed for surgical automation.

\subsection{Binary Classifier for Visual Data}
The system processes pixel data from microscope and B-scan images to make binary decisions at different procedural stages. This defines a mapping from pixel-level visual data to binary outcomes, such as contact/no contact and puncture/no puncture.  The system uses visual information from the images to make decisions for needle navigation, contact detection, and puncture detection.

We can use  \( D = \{(I_i, b_i)\}_{i=1}^N \) to represent pairs of input and their corresponding labels, where \( I_i \) represents the pixel data from the input images (either microscope or B-scan), and \( b_i \in \{0, 1\} \) denotes the binary decision (e.g., puncture = 1, no puncture = 0, contact = 1, no contact = 0), the objective is to construct a mapping that maps the input pixel data to a binary decision.
The system follows this sequence of steps:

\noindent \textbf{Needle Tip Detection and Navigation}: A deep learning model detects the needle tip and calculates its movement toward a predefined 2D target within microscope images. The needle continues navigating until its Euclidean distance from the target is below 3 pixels.

\noindent \textbf{Contact Detection}: For contact detection, the ground truth was defined by expert annotation of B-scan frames showing the first continuous contact between the needle and the vein wall, as explained in Section III.

Once the needle reaches the target, the system switches to processing B-scan images to determine whether contact with the vessel has been made. A binary decision \( b \in \{0, 1\} \) is produced: \( b = 1 \) indicates contact, and \( b = 0 \) indicates no contact. If no contact is detected, the needle's position is adjusted along the negative z-axis to get closer to the vein until contact is confirmed.

\noindent \textbf{Puncture Detection}: Once contact is confirmed, the system advances to the puncture phase, increasing velocity to perform the puncture. B-scan images are analyzed to determine the success of the puncture. The final binary decision \( b = 1 \) (puncture) or \( b = 0 \) (no puncture) determines the outcome. If no puncture is detected, the system adjusts and retries until a successful puncture is achieved.
\section{METHODS} 
The system illustrated in Fig. \ref{fig:1} integrates two robotic platforms: EyeRobot 2.0~\cite{uneri2010new} performs needle-based vein puncture, while EyeRobot 2.1~\cite{he2012toward} uses a silicone vein holder to stabilize the target vessel (Fig. \ref{fig:1}(b)). The OCT system provides real-time imaging, capturing both microscope and B-scan images. The proposed workflow is illustrated in Fig. \ref{fig:2}. A user-defined target is selected on the microscope image. The navigation model detects the needle tip and computes a velocity vector toward the target. The robot iteratively adjusts movement until the Euclidean distance between the needle tip and target is below 3 pixels (0.0586 mm/pixel at the ROI). Upon arrival, the B-scan images are fed into a contact detection model. If no contact is detected, the needle is moved downward along the robot z-axis until contact is confirmed.
 (see Fig. \ref{fig:1}(a)) Following that, the puncture detection model is engaged. If a puncture is not detected, the needle retracts by 2/5 of the puncture depth before reattempting, a strategy based on experiments to minimize needle stress while preventing withdrawal from the vein. This process iterates until a successful puncture is achieved, after which the needle is fully retracted, completing the procedure.
\subsection{Needle Tip Detection and Navigation}
Detecting the needle tip during OCT-guided puncture involves precisely tracking it as it interacts with delicate tissues. Microscope-based needle tracking is challenging when vessels occlude the needle tip, reducing visibility and increasing detection errors. Conventional image-processing techniques struggle under these conditions. The needle tip detection network addresses this by generating a fixed bounding box around the ROI. We use a ResNet-101 backbone with a key point RCNN head, followed by a Feature Pyramid Network (FPN) to extract and aggregate the features from the microscope images. We selected ResNet-101 and FPN for their multi-scale feature representation, enabling needle tip detection even under partial occlusion. The RCNN structure offers interpretable keypoint detection in the microscope image. We chose a learned approach instead of direct coordinate transformation or visual servoing. The needle tip is often occluded by the vessel wall or tissue, which may reduce the accuracy of geometric model-based tracking. We chose this data-driven approach over coordinate transformation or visual servoing for two reasons. First, due to tissue deformation and imaging variability, it is difficult to define consistent geometric mappings between image and robot space. Second, the needle tip is often occluded or poorly contrasted, limiting the reliability of geometry-based tracking. By learning directly from image data, the model adapts to these visual variations and provides more stable predictions in real time.

 The Region Proposal Network (RPN) generates proposals for regions likely to contain the objects of interest. The ROI heads refine these proposals and perform final object classification, bounding box regression, and key point detection. After the needle tip is detected, the navigation module generates a velocity vector that connects the detected needle tip to the user-defined target. Note that the robot is manually leveled so that the XOY plane of the task space aligns with the microscope's image plane. This ensures the robot moves along the horizontal plane. The navigation stops when the pixel difference is smaller than a specific criterion (empirically set to 3 pixels). 

\subsection{Contact Detection}
The contact detection network determines whether the needle makes contact with the vein during OCT-guided procedures. It uses the ResNet-18 architecture, pre-trained on ImageNet and adapted for binary classification.
 The final fully connected layer is replaced with a two-node output layer corresponding to two classes: contact (1) or no contact (0). This architecture extracts features from B-scan images to classify contact events.
During training, the model uses a supervised approach with labeled B-scan images. These images are processed through the ResNet-18 backbone, extracting features via convolutional layers. A final fully connected layer outputs class probabilities for the two classes. The model is optimized using a cross-entropy loss function and trained with stochastic gradient descent, using a learning rate of 0.001 and a momentum of 0.9. The ground truth contact labels were manually annotated based on expert observation of B-scan sequences, defining contact as the first sustained visual overlap between needle and vessel wall. Although the physical contact may be gradual, we binarized the contact state for control simplicity and consistency in real-time inference.

 Data augmentation techniques such as horizontal flipping and normalization are applied to improve model generalization.
In the inference phase, the trained model receives a B-scan image with a size of 224$\times$224 pixels (0.0357 mm/pixel at the ROI). 
The network outputs class probabilities.
If the probability of the contact class exceeds a predefined threshold, the system predicts contact; otherwise, it predicts no contact. This real-time inference provides iterative feedback, allowing the system to adjust the needle position along the z-axis until contact is confirmed.

\subsection{Puncture Detection}
The puncture detection network determines whether the needle has successfully punctured the vein during OCT-guided procedures.
 The model is based on the YOLOv8x architecture, known for its real-time detection capabilities. This architecture uses a CNN backbone to extract deep features from input B-scan images. These features are then processed into a grid of cells for object detection and classification.
Each cell predicts multiple bounding boxes with a confidence score indicating whether it contains the needle tip.
 Additionally, the network simultaneously classifies whether a puncture has occurred by predicting a binary class label for each bounding box.
The model is optimized using a combination of binary cross-entropy loss for classification and regression loss for bounding box prediction.
During inference, the model uses live-streamed B-scan images to estimate (1) the bounding box coordinate around the needle tip, (2) a binary label for puncture, and (3) the confidence score for the bounding box prediction.

If no puncture is detected with high confidence, the system retracts and reattempts the puncture. This process is repeated until the model confirms a successful puncture.

\section{Experiment}
\subsection{Experimental Setup}
Our experimental setup consists of two robotic systems (EyeRobot~2.0 and EyeRobot~2.1), a surgical needle with a 3D-printed tool holder, a customized vein holder, a chicken embryo holder, and a customized infusion system with two syringes connected by a tube. An intraoperative OCT system (Leica Proveo~8 with EnFocus OCT) provides a light source and captures both microscope and B-scan images throughout the procedure. Fig. \ref{fig:1} illustrates the experimental layout. EyeRobot 2.0 is responsible for needle insertion using a MedOne MicroTip Beveled Cannula (25g/40g) with a 100-micron tip, positioned at a 70-degree angle for precise puncture of the chicken embryo veins. EyeRobot 2.1 serves as a stabilizer, securing and elevating the target vein during the procedure. The vein holder material provides structural support while allowing multiple punctures without damaging the needle. Before running the experiment, the top membrane of the chicken embryo was removed using forceps (STORZ Castroviejo Suturing Forceps E1796) to expose the veins for the procedure. During the experiments, the robots are controlled via motorized joints, and the tooltip positions are tracked based on the robot's kinematics. The needle was initially fixed at a 70-degree angle relative to the vein plane. Once fixed, the needle was moved along this direction via keyboard control to approach or depart from the vein, depending on the procedure stage. For the puncture, the robot advanced the needle into the vein at high velocity.

\begin{figure*}[tb]
    \centering
    \includegraphics[width=0.9\textwidth]{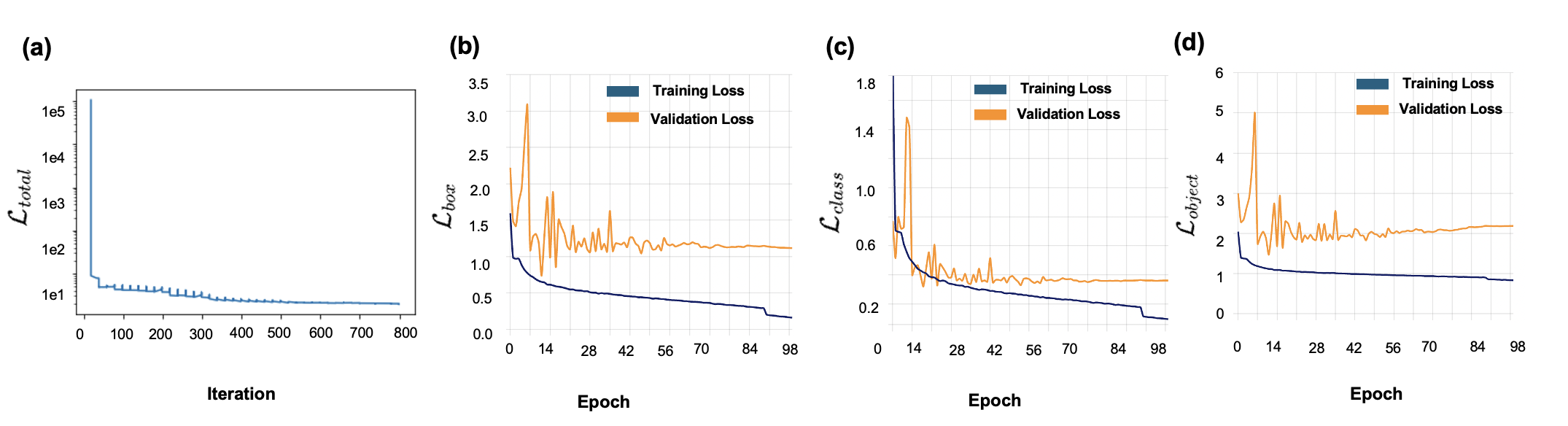}  
    \caption{Training results of navigation model and puncture detection model: total loss (a) of key point detection model (Detectron2 - Mask RCNN), box loss (b), class loss (c), and object loss (d).}
    \label{fig:3}
    \vspace{-16pt}
\end{figure*}

\subsection{Training }
\noindent \textbf{Data Collection: }
Data from manual needle injections on 30 chicken embryos were collected, generating a total of 47,342 images, including both microscope and B-scan images. The average vein diameter was 1.27 mm. From this dataset, 17,000 microscope images were selected for training the needle navigation model, 4,353 B-scan images for contact detection training, and 4,730 B-scan images for puncture detection training.
Throughout the procedure, microscope and B-scan images were recorded by the OCT system to capture needle-vein interactions.\\
\noindent\textbf{Data Augmentation:}
To ensure robust learning for our models, we employed brightness adjustment and Gaussian blur on the microscope images. The B-scan dataset was expanded to 11,350 images through various augmentation methods, including brightness adjustment (ranging between -15\% and +15\%), exposure modification (between -10\% and +10\%), and noise addition (up to 0.1\% of pixels). Each image was annotated with a bounding box around the needle tip. These augmentation techniques helped the network generalize to unseen conditions and improved its robustness in varied environments.\\
\noindent\textbf{Networks Training:}
The navigation model was trained using the Detectron2 framework to detect the needle tip as a key point within the microscope images. Each image is labeled with the needle tip as a key point and a $50\times50$ pixel bounding box around the key point. The training was conducted with a batch size of 16 and an initial learning rate of 0.001. The cross-entropy loss was utilized for key point detection, and the model was trained for 4,500 iterations.
The loss curve demonstrates rapid convergence, with the key point loss stabilizing around 1.7351 (see Fig. \ref{fig:3}(a)). This indicates that the model effectively learned the features necessary for precise needle tip detection in microscope images showing minimal overfitting.
The contact detection model was implemented on a ResNet-18 backbone, pre-trained on ImageNet, and fine-tuned for binary classification to detect needle-vein contact in B-scan images. The final fully connected layer was replaced with a two-node output layer corresponding to the classes of contact and no contact. Training was conducted for 100 epochs.
The model achieved 98\% accuracy, evaluated with standard classification metrics on a test set of 200 B-scan images. 
We trained a YOLOv8x for puncture detection, which achieved a mAP50 (mean average precision with IOU over 50) of nearly 0.96 and an mAP50-95 of 0.78, which indicates high precision and recall.
The training process showed a steady reduction in box loss, class loss, and object loss, as illustrated in Fig. \ref{fig:3}(b), \ref{fig:3}(c), and \ref{fig:3}(d). Box loss measures the accuracy of the predicted bounding box, class loss evaluates the correctness of the puncture classification, and object loss reflects the model's confidence in detecting the puncture, with lower values indicating better performance in each aspect. While training loss decreased consistently, validation loss exhibited fluctuations, with higher variance in object loss. This discrepancy suggests increased complexity in the validation dataset. Despite this, the model maintained strong performance in real-time puncture detection, demonstrating reliability for autonomous surgical procedures.

\subsection{Validation method}
\begin{figure}[tb]
    \centering
    \includegraphics[width=\linewidth]{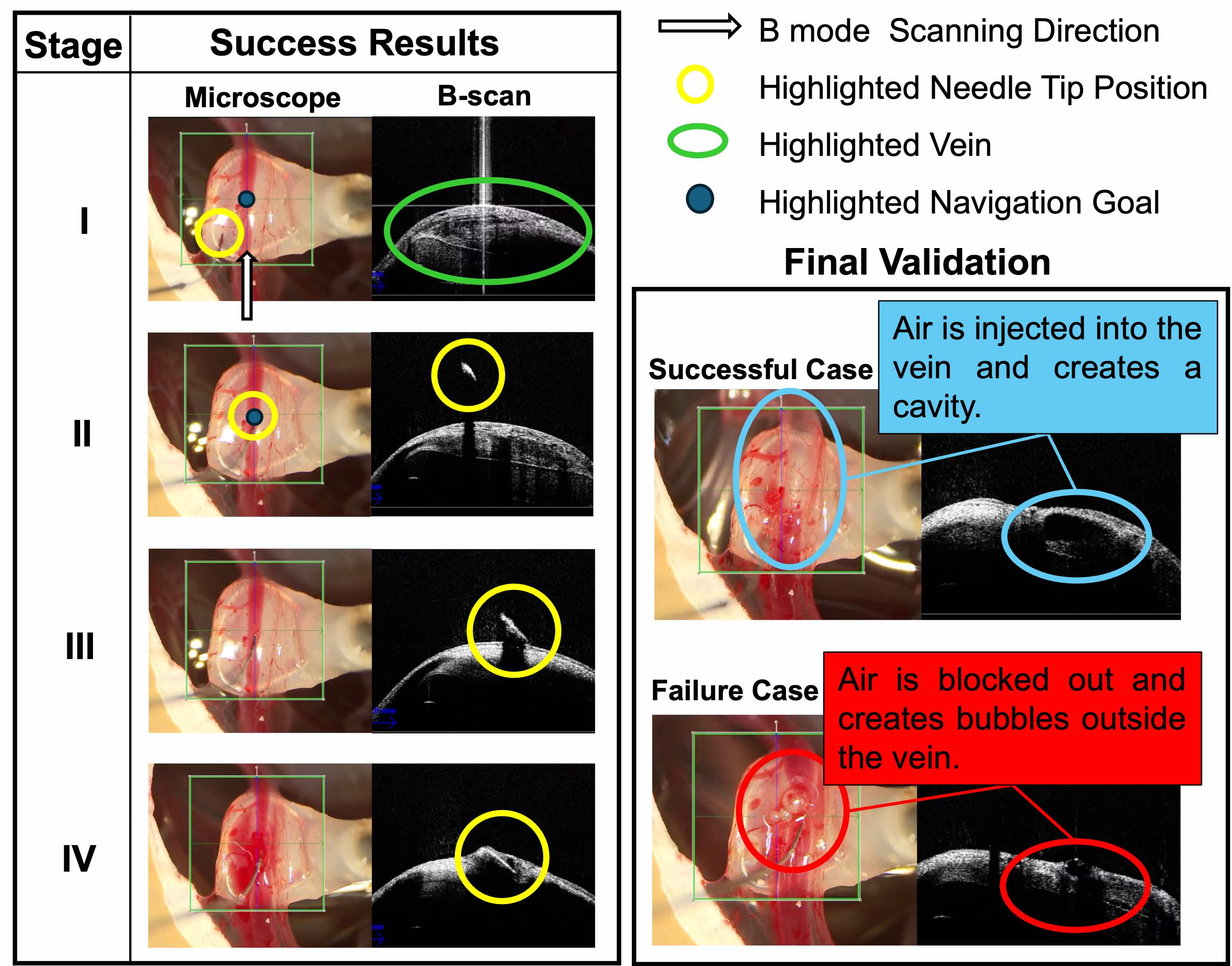}
    \caption{Exemplary results of every module and the final validation result. The stages are denoted as I: Initial Stage; II: Needle Navigation Stage; III: Contact Detection Stage; IV: Puncture Detection Stage. The end result is validated by pumping air through the needle. Stage II uses microscopic image as the primary view; Stage III and IV use B scan image as the primary view. In stage II, the possible failure case would be when the needle is not driven close enough to the manually selected navigation goal. An incorrect outcome is identified by checking whether the air is injected into the vein. 
}   
    \label{fig:4}
    \vspace{-16pt}
\end{figure}

To assess the performance of the navigation, contact detection, and puncture models, we implemented a structured validation process replicating realistic surgical conditions in live chicken embryos.
 The validation began with the navigation model, where the needle was guided toward a user-defined target by iteratively adjusting its movement based on the model’s predictions. The model refined the needle's position until the error between the predicted needle tip location and the target was reduced to less than 3 pixels (0.0586 mm/pixel at the ROI). This iterative approach ensured precise navigation within the complex vascular environment of the chicken embryo. The navigation model accuracy was measured by the pixel distance between the final needle position and the target, with success defined as within 3-pixel threshold.
The contact detection module is activated once the needle reaches the target.  
If no contact is detected, the system adjusts the needle position along the negative z-axis in robot space, continuously re-evaluating the contact status using the contact detection model.  The process is repeated until the model confirms contact, indicating that the needle is correctly positioned against the vein.
Following successful contact detection, the puncture module was validated by injecting air through the needle into the target vein. 
 The success of the puncture is confirmed by observing visible inflation of the vein and blood exiting from the puncture site.  If neither inflation nor blood was detected, the experiment was marked as a failure, indicating an incorrect puncture prediction by the model (see Fig. \ref{fig:4}). 
This multi-stage validation procedure was conducted across 20 embryos, ensuring that the models could consistently perform accurate navigation, contact detection, and puncture detection under dynamic, real-world conditions.
In some trials, failures occurred due to lighting variation or tissue motion causing poor contrast in B-scan images, which affected needle localization. These failure cases are rare but suggest that further spatial accuracy analysis, including depth control precision, would be important in future work.

\section{Result and Discussion}

\begin{figure}[tb]
    \centering
    \includegraphics[width=\linewidth]{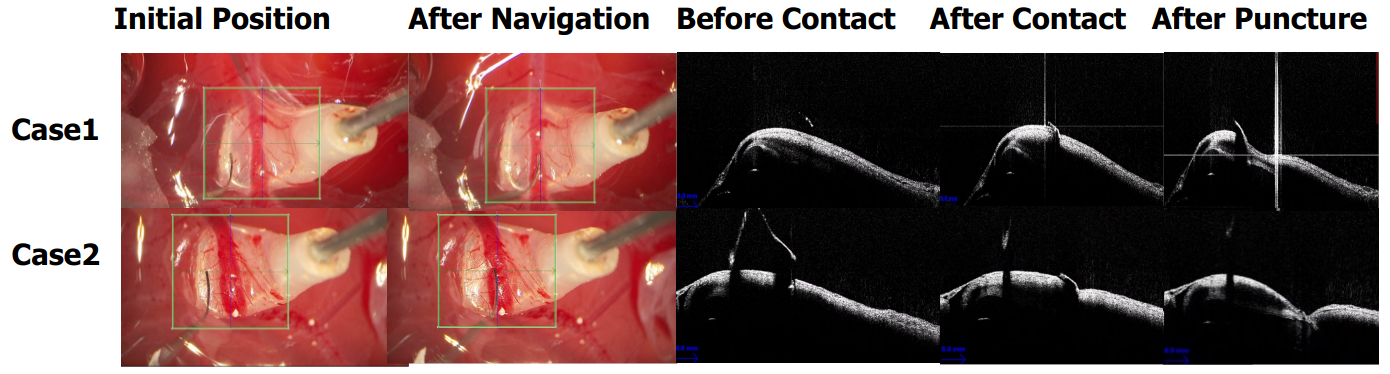}  
    \caption{Representative examples from the validation experiments for retinal vein cannulation. Out of 20 validation trials, this figure presents two cases where the model successfully detected the puncture outcome. The top row shows a failed puncture attempt, which the model correctly identified as unsuccessful. The bottom row shows a successful puncture, which the model also accurately classified as a success. Each column represents different procedural stages: initial position, post-navigation, pre-contact, post-contact, and post-puncture. }
    \label{fig:5}
    \vspace{-12pt}
\end{figure}

Experiments on 20 chicken embryos were conducted to assess the performance of the autonomous system. Each trial tested the system's ability to navigate to the target and perform the puncture accurately. The results of these trials, including both successful and failed puncture attempts, are illustrated in Fig. \ref{fig:5}, where the system's ability to differentiate between successful and unsuccessful cannulations is demonstrated.

\begin{table}[htbp]
\caption{Average, Median, and Standard Deviation of Navigation and Puncture Time.}
\vspace{-8pt}
\label{tab:Time_table}
\begin{center}
\begin{tabular}{c c c}
\hline
\rowcolor{red!10!blue!15}
\multicolumn{1}{p{2.6cm}}{\centering Metric}
& \multicolumn{1}{p{2.0cm}}{\centering Navigation Time (seconds)}
& \multicolumn{1}{p{2.0cm}}{\centering Puncture Time (seconds)}
\\
\hline
Average & 36.74 & 26.97 \\
Median & 35.87 & 24.13 \\
Standard Deviation & 12.73 & 13.68 \\
\hline
\end{tabular}
\end{center}
\vspace{-12pt}
\end{table}

\begin{table}[ht]
\caption{Puncture Model Performance in Real World Experiments.}
\vspace{-8pt}
\label{tab:Precision_table}
\begin{center}
\begin{tabular}{c c c}
\hline
\rowcolor{red!10!blue!15}
\multicolumn{1}{p{2.3cm}}{\centering Metric}
& \multicolumn{1}{p{2.3cm}}{\centering Class 0 (Failure)}
& \multicolumn{1}{p{2.3cm}}{\centering Class 1 (Success)}
\\
\hline
Precision & 0.9 & 0.82 \\
Recall & 0.75 & 0.93 \\
F1-score & 0.82 & 0.87 \\
Support & 12 & 15 \\
\hline
\end{tabular}
\end{center}
\vspace{-12pt}
\end{table}

\begin{figure}[tb]
    \centering
    \includegraphics[width=\linewidth]{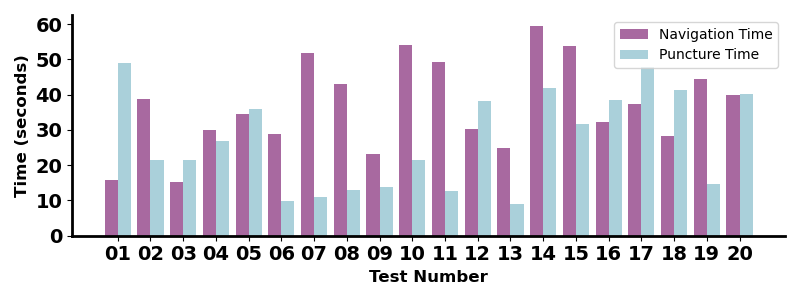}  
    \caption{The navigation and puncture time for 20 test trials. Each test number corresponds to an individual experiment, with navigation time (purple) and puncture time (blue) measured in seconds. The results highlight variations in task duration across trials.}
    \label{fig:6}
    \vspace{-16pt}
\end{figure}

In each experiment, needle navigation was first performed manually by an expert user with the microscope and EyeRobots, followed by autonomous navigation. As shown in Fig. \ref{fig:6}, the autonomous system improved navigation time to 36.74 s on average.
This shows a 68.6\% reduction in navigation time compared to the manual method (117 s on average). Fig. \ref{fig:6} also shows that
the automated puncture system improved puncture time to an average of 26.97 s,significantly faster than the 469 s in manual procedures.

 However, the standard deviation of 12.73 secs in navigation, and the standard deviation of 13.68 secs in puncture(see Table \ref{tab:Time_table})  indicate that the proposed networks' performance varied across trials, likely due to differences in vein positioning, lighting conditions, and embryo movement.  One major factor was lighting variation in OCT images, which affected contrast and visibility, sometimes leading to incorrect needle tip localization. These contrast changes and occlusions occasionally caused longer navigation times, as the system required additional adjustments to correctly detect and align the needle tip. More importantly, three outlier cases were observed with an average navigation time of 214.84 s. In these cases, needle occlusion by vessels or membrane layers reduced the ability of the model to track the needle accurately. Certain puncture trials encountered difficulties, possibly due to slight misalignments or tissue interactions during the process. Puncture time variations 
 could result from differences in vein stiffness and the contact detection model's sensitivity to subtle differences in tissue interaction. 

It is important to note that the “manual method” in this study refers to controlling the robotic system via a keyboard by an expert user.
 While this provides a baseline for comparison within our experimental setup, it differs from traditional RVC performed entirely by a surgeon without robotic assistance. 

The system achieved an 85\% classification accuracy for puncture detection. Precision was 0.90 for Class 0 (Failure) and 0.82 for Class 1 (Success), while recall values were 0.75 and 0.93, respectively. The model demonstrated strong sensitivity in detecting successful punctures, though challenges remain in minimizing misclassified failure cases. These metrics are reflected in the F1-scores of 0.82 for failures and 0.87 for successes shown in Table \ref{tab:Precision_table}. Compared to prior studies, Willekens et al.~\cite{willekens2017robot} reported a 73\% success rate in robot-assisted retinal vein cannulation, defining complete success as maintaining cannulation for more than three minutes.
This accuracy suggests the system is reliable in real time, though improvement is needed to reduce false positives.  In addition, the current vision pipeline is sensitive to domain shifts, including changes in tissue appearance, OCT contrast, and illumination.
 Needle tracking accuracy may degrade under these conditions, which could affect navigation or puncture detection performance. These limitations should be addressed by extending the dataset or incorporating more robust feature representations in the future.


\section{Conclusion} \label{sec:conclusion}
This paper presented an autonomous robotic system for precise retinal vein cannulation using OCT imaging and deep learning. Our approach uses a deep learning-based navigation model for real-time needle tip detection, with contact and puncture detection modules.
 By incorporating microscope and B-scan images, the system could navigate the surgical environment and perform the puncture reliably.

Validation on live chicken embryo models demonstrated a 68.6\% reduction in navigation time and 85\% classification accuracy for puncture detection, indicating that deep learning can support precise, real-time decision-making for RVC.

Future work could combine contact and puncture detection into a unified model to streamline decision-making.
Adaptive image preprocessing, such as contrast normalization and denoising, could improve consistency under varying conditions, reducing the impact of lighting variations and needle occlusions. Implementing a confidence-based decision-making process could further improve reliability by allowing the system to proceed with puncture only when the model surpasses a confidence threshold. Additionally, a retry mechanism could help the system adjust the needle position when uncertainty arises.

Furthermore, because the chicken embryo veins are larger than human retinal veins, future studies should validate the system in \textit{ex vivo} porcine eyes with smaller vessel diameters and more similar vascular structures.

These advancements could refine surgical precision and expand applicability to complex human retinal conditions. Further validation on porcine or human retinal samples will be necessary to evaluate performance under more realistic anatomical and optical conditions. In addition, the current models are trained in a supervised fashion using a single dataset; their robustness under domain shifts remains limited. Future work could explore foundation models, such as vision transformers pre-trained on large-scale medical data, or integrate geometry- or physics-based priors to improve generalization. Moreover, we did not quantify the needle’s depth control accuracy nor perform a systematic analysis of failure modes. These remain important directions for future work to ensure spatial precision and operational reliability.

\addtolength{\textheight}{-7.9cm}   

\bibliographystyle{IEEEtran}
\bibliography{bibliography}

\end{document}